\documentclass[10pt,twocolumn,letterpaper]{article}

\usepackage{iccv}
\usepackage{times}
\usepackage{epsfig}
\usepackage{graphicx}
\usepackage{amsmath}
\usepackage{amssymb}
\usepackage{booktabs}
\usepackage[accsupp]{axessibility}

\newcommand*{\argmax}{\textrm{argmax}}

\usepackage[pagebackref=true,breaklinks=true,letterpaper=true,colorlinks,bookmarks=false]{hyperref}

\iccvfinalcopy 

\def\iccvPaperID{8} 

\ificcvfinal\fi

\begin{document}

\title{Latent Variable Models for Visual Question Answering}

\author{Zixu Wang$^{1}$, Yishu Miao$^{1}$, Lucia Specia$^{1,2}$ \\
$^1$Department of Computing, Imperial College London, UK\\
$^2$Department of Computer Science, University of Sheffield, UK\\
{\tt\small \{zixu.wang, y.miao20, l.specia\}@imperial.ac.uk}
}

\maketitle
\ificcvfinal\thispagestyle{empty}\fi

\begin{abstract}
Current work on Visual Question Answering (VQA) explore deterministic approaches conditioned on various types of image and question features.
We posit that, in addition to image and question pairs, other modalities are useful for teaching machine to carry out question answering.
Hence in this paper, we propose latent variable models for VQA where extra information (\eg captions and answer categories) are incorporated as latent variables, which are observed during training but in turn benefit question-answering performance at test time.
Experiments on the VQA v2.0 benchmarking dataset demonstrate the effectiveness of our proposed models: they improve over strong baselines, especially those that do not rely on extensive language-vision pre-training.
\end{abstract}

\section{Introduction}
As a classic multi-modal machine learning problem, Visual Question Answering (VQA)~\cite{VQA} systems are tasked with providing a correct textual answer given an image and a textual question.
Current VQA models~\cite{shih2016wtl, kazemi2017show, anderson2018bottom} are trained to learn the relationship between areas in an image and the question, and to choose the correct answer from a vocabulary of answer candidates, \ie, they are modelled as a classification problem.
The majorities of popular VQA~\cite{goyal2017making, NIPS2018_7429, yu2019mcan} models are created in a deterministic manner and explore solely information from the given image-question pair.
There are other approaches attempting to incorporate extra information, such as image captions~\cite{wu-etal-2019-generating} and mutated inputs~\cite{gokhale2020mutant, Chen_2020_CVPR}.
However, it in turn restricts the practical applications as the extra information is required to be explicitly available during testing.  

In this paper, we propose an approach to explore additional information as latent variables in VQA: we employ latent variables for VQA to exploit extra information (\ie image captions and answer categories) to complement limited textual information from image and question pairs. 
We assume a realistic setting where this information -- esp. captions -- may only be available during the training phase.
To that end, we introduce a continuous latent variable as the caption representation to capture the essential information from this modality.
Moreover, the answer category is modelled as a discrete latent variable, which acts as an inductive bias to benefit the learning of answer prediction, and can be integrated out during testing. 
The motivation is that the generative framework is able to incorporate many other types of information as continuous or discrete latent variables, and as such it effectively leverages additional resources to constrain the original image-question distribution while omitting them in testing.
This grants the models with stronger generalisation ability compared to its deterministic counterparts, which generally require off-the-shelf pipelines to model the information from external modalities.

Intuitively, image captions describe diverse aspects of an image and include attributes and relations of objects in a more informative way. 
In our work, a continuous latent variable is employed for capturing the caption distributions and constraining the generative distribution conditioned on image and question pairs.
In this way, the joint multimodal representations from images and question can benefit from the caption modality during training, and it requires no explicit caption inputs in testing.
Similarly, there exists a strong connection between a question and answer pair when the question provides informative signals on its type or the category of possible answers. 
For example, ``How many", ``Where is" and ``what is" normally connect to numbers, locations, and objects respectively. 
We propose a discrete latent variable is employed for modelling answer categories and providing better inductive bias from the question and answer pairs.

In summary, our {\bf main contributions} are:
\begin{itemize}
    \item A novel generative VQA framework combining the modularity of latent variables with the flexibility to introduce extra information as continuous and/or discrete latent variables.
    \item A method to incorporate additional information which does not rely on building multiple deterministic pipelines, aiming at learning the underlying compositional, relational, and hierarchical structures of multiple modalities. The models benefit from the extra information during training without providing explicit inputs in testing. 
    \item The improvements over deterministic baseline models (\eg UpDn \cite{anderson2018bottom} and VL-BERT \cite{Su2020VL-BERT}) in experiments with the VQA v2.0 dataset demonstrate the effectiveness of our proposed latent variable models.
    Our qualitative analysis also indicates that using extra resources (\ie captions and answer categories) as latent variables captures complementary information during training and benefits the VQA performance in testing.
\end{itemize}

\begin{figure}[!t]
  \centering
  \includegraphics[width=0.5\textwidth]{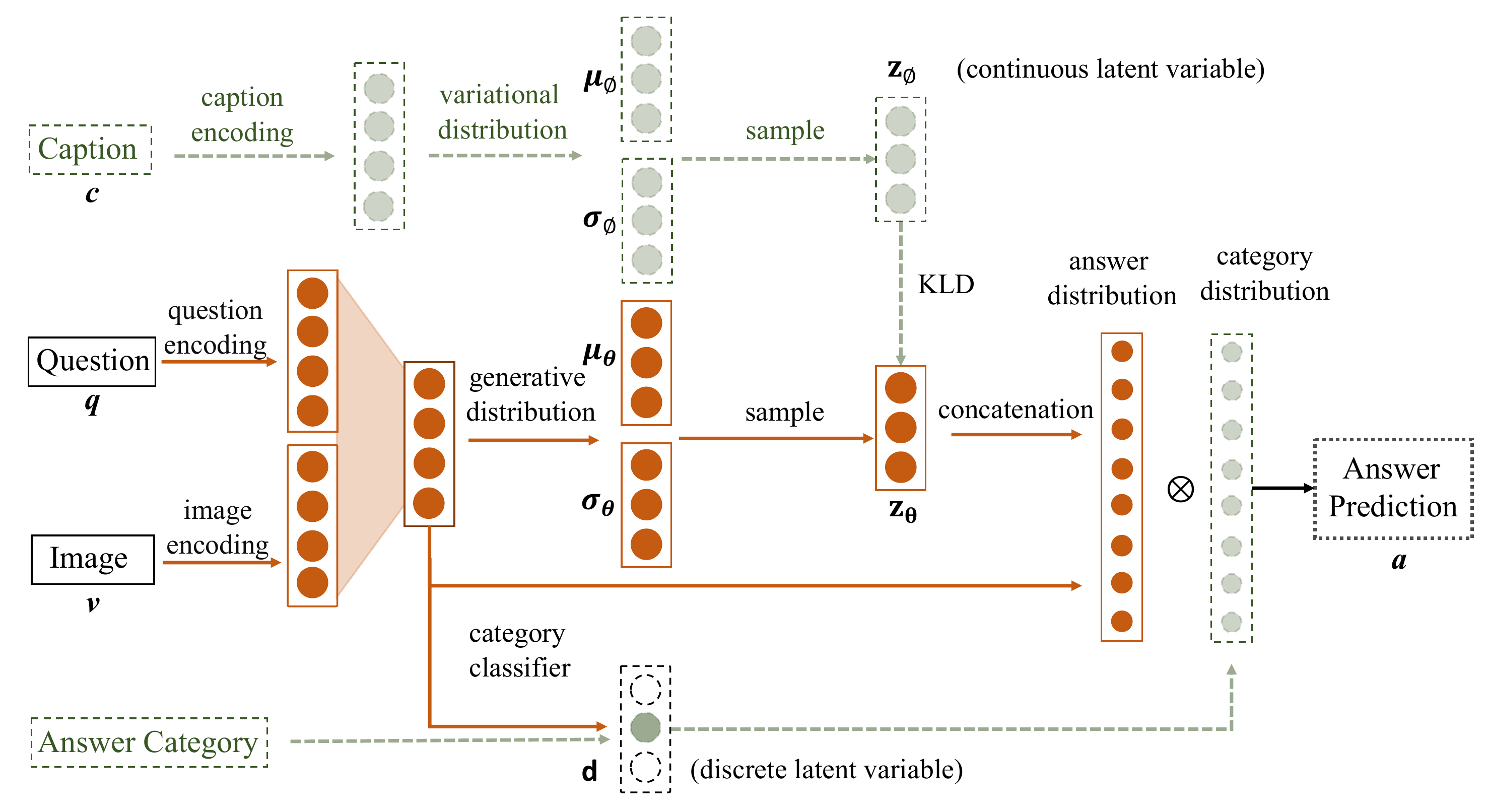}
  \caption{Architecture of our latent variable model for VQA. We use dotted lines to denote the process of proposed latent variables. 
  }
  \label{fig:metho}
\end{figure}

\section{Model}
We first present an overview of our general model structure, followed by the encoders for different modalities, and the proposed corresponding latent variables. 

\subsection{General Model Structure}
\label{sec:general_struc}
In a VQA task, images and questions are normally used to learn a joint multimodal distribution for answer predictions.
We postulate that the joint representation can be improved by other multimodal information.
Hence, we introduce captions and answer categories to our VQA model as continuous and discrete latent variables respectively to encourage a better learning in the joint distribution of image and question pairs during training.
A notable advantage of the latent variable models is that they do not explicitly require captions or answer categories during testing, and therefore can be easily extended to condition on any other useful information.

Firstly we introduce the notations used in the general VQA model. $V$, $Q$, $A$ are used to denote the input image, question, and answer instances respectively.
The image feature $v$, question representation $q$, and answer representation $a$ are extracted from the image encoder, question encoder, and answer encoder.
The VQA task is constructed as a classification problem to output the most likely answer $\hat{a}$ from a fixed set of answers based on the content of the image $v$ and question $q$: 
\begin{equation}
    \hat{a} = \argmax \ p(a|v, q)
\end{equation}

In our latent variable model, we introduce image captions $C$ to the training phase.  
Similarly, we extract the caption features $c$ by a caption encoder. 
However, instead of directly feeding in the caption features $c$ into to the model, we employ a continuous latent distribution $z$ to be the caption representations.
Here $z\sim q(z|c)$ is modelled as variational distribution. 
We then build a generative distribution $z\sim p(z|v,q)$ to infer the caption information by conditioning on image and question pairs, which is optimised during training via neural variational inference. 
We originally experimented using $q(z|v, q, c)$ as the vairational distribution. 
However, this distribution is quite close (\ie small KL divergence) to the generative distribution $p(z|v,q)$, which weakens the learning signal from KL divergence.

In addition, we introduce a discrete latent variable $d$ for modelling answer category inferred via $d\sim p(d|v,q)$, which is also conditioned on image and question pairs. 

Hence, the training of the latent variable model is carried out by the samples $(v, q, a, c, d)$. During testing, the answer $a$ is predicted from the image and question pair $(v, q)$:
\begin{equation}
    \hat{a} \!=\! \argmax \!\sum_{d,z} p(a|v, q, d, z) p(d|v, q) p(z|v, q)
\end{equation}
where the discrete latent variable $d$ is directly integrated out, and the $z$ is the Monte-Carlo sample from $p(z|v,q)$.

\subsection{Continuous Latent Variable: Caption}\label{subsec:cap_enc}
As captions are modelled by a continuous latent variable, we only have explicit captions during training. 
Here we present the generative distribution that is conditioned on images and questions during testing, and the variational distribution that is conditioned on explicit captions during training.
Therefore, the caption encoder is only used in the training phase. 

\noindent\textbf{Generative Distribution - $p_{\theta}(z|v, q)$}. We use a latent distribution $p_{\theta}(z|v, q)$ to model the joint multimodal distributions of images and questions. Compared to its deterministic counterpart using concatenated multimodal features, we parameterise the stochastic distribution with $\mathcal{N}(z|\mu_{\theta}(v, q), \sigma^2_{\theta}(v, q))$. 

\noindent\textbf{Variational Distribution - $q_{\phi}(z|c)$}.
We first apply a RNN model to embed the caption inputs $C$ and a latent variable $q_{\phi}(z|c)$ to model the caption semantics and distributions, where $z \sim \mathcal{N}(z|\mu_{\phi}(c), \sigma^2_{\phi}(c))$. 

\subsection{Discrete Latent Variable: Answer Category} \label{subsec:answer_cat}
Assuming that each image and question pair $(v, q)$ can be projected to an answer category to help find a correct answer, we are able to encourage the model to distinguishing candidates across answer categories instead of only the spurious relationships between questions and answers via simple linguistic features.
Therefore, in order to leverage this useful inductive bias, we propose a discrete latent variable to model the answer category given an image and question pair $(v, q)$.
In particular, for each answer category $d$, we have a conditional independent distribution $p(a|v, q, d)$ over the answers in the certain answer category. 
\begin{equation}
    p(a|v, q) =  \sum_d p(a|v, q, d) \cdot p(d|v, q)
\end{equation}
We trained an answer category classifier using joint image-question pairs as the input, given the true labels as shown at left bottom in Figure \ref{fig:metho}. We then use the category distribution to modify the answer distribution through element-wise production to get more precise answer distribution.

\section{Datasets \& Setup}
\subsection{Datasets}
\label{sec:data}
We use the VQA v2.0 dataset
~\cite{VQA} for our proposed latent variable model. 
The answers are balanced in order to minimise the effectiveness of dataset priors. 
We report the results on validation set and test-standard set through the official evaluation server. 
The source of image captions in our work is the MSCOCO dataset
~\cite{lin2014microsoft}. 

We use answer categories from the annotations of ~\cite{krishna2019information}. 
The answers in the VQA v2.0 dataset are annotated with a set of 15 categories for the top 500 answers that makes up the 82\% \footnote{Although the category definitions cannot cover all types of answers, and false prediction during testing might be observed, the latent variable can still maintain the robustness in predicting correct answers by summing over all the probabilities of predicted categories.} of the VQA v2.0 dataset; and the other answers are treated as an additional category. 

\section{Experiments}

In this section, we first describe the experimental results of our latent variable model, compared with both a UpDn (Bottom-up Top-down) baseline model and a state-of-the-art pre-trained visual-linguistic model (VL-BERT); then we conduct qualitative analysis to validate the effectiveness of proposed components.

\begin{table}[!t]
\centering
\addtolength{\tabcolsep}{-1.7pt}
\scalebox{0.9}{
\begin{tabular}{l|cccc|c}
\hline 
\toprule
& \multicolumn{4}{|c|}{VQA v2.0 test-dev (\%)} & test-std (\%) \\ 
\hline
& {\footnotesize All} & {\footnotesize Yes/No} &  {\footnotesize Num} & {\footnotesize Other} & {\footnotesize All} \\ 
\hline
\hline
Caption~\cite{wu-etal-2019-generating} & - & - & - & - & 68.37 \\
DFAF~\cite{peng2018dynamic} & 70.22 & 86.09 & 53.32 & 60.49 & 70.34\\
MLIN~\cite{Gao_2019_ICCV} & 71.09 & 87.07 & 53.39 & 60.49 & 71.27 \\
\midrule
UpDn~\cite{anderson2018bottom} & 65.32 & 81.82 & 44.21 &  56.05 & 65.67 \\
+ latent (ours) & \underline{66.01} &  82.96 &  44.58 & 55.94 & \underline{66.29} \\
\midrule
VL-BERT$_{\textrm{large}}$~\cite{Su2020VL-BERT} & 71.79 & - & - & - & 72.22 \\
+ latent (ours) & \bf \underline{72.03} & 88.03 & 54.16 & 62.42 & \bf \underline{72.37} \\ 
\bottomrule
\end{tabular}}
\caption{Experimental results on VQA v2.0 test-dev and test-standard (test-std) set. Accuracies are reported in percentage (\%) terms. The state-of-the-art scores are in bold; underlined scores are best among (baseline \vs latent variable extension); and both underlines and bold scores are the overall best results.}
\label{tab:vqa_results}
\end{table}

\subsection{Quantitative Analysis}
We compare the results of our latent variable model with the baseline model (UpDn), a state-of-the-art visual-linguistic pre-training model (VL-BERT), and three other related VQA models; where~\cite{wu-etal-2019-generating} uses generated captions to assist answer predictions and~\cite{peng2018dynamic, Gao_2019_ICCV} explore the interactions between visual and linguistic inputs. 

As demonstrated in Table~\ref{tab:vqa_results}, our latent variable model outperforms when acting as an extension. 
In particular, our latent variable model outperforms UpDn by 0.69\% accuracy on test-dev set and by 0.62\% accuracy on test-standard set. 
In addition, our model improves the performance by 0.24\% accuracy than its VL-BERT counterpart on test-dev set and by 0.15\% accuracy on test-standard set. 
These results indicate the effectiveness of including captions and answer categories as latent variables, to promote the distribution of image and caption pairs to be closer to the captions' space, and to learn a better distinction among different kinds of answers, or different answers within the same answer category. 

The result of~\cite{wu-etal-2019-generating} (68.37\%) is a very strong baseline, which follows a traditional deterministic approach. 
However, their model is trained to generate captions that can be used at test time, while in our case only image and question pairs are required for answer prediction. 
\cite{peng2018dynamic} and \cite{Gao_2019_ICCV} both achieve comparable performance (70.34\% and 71.27\% on test-standard set, respectively) to VL-BERT (72.22\%) without pre-training, by dynamically modulating the intra-modality information and exploring the latent interaction between modalities. 
Our latent variable model has the overall best result when combined with the strong pre-training VL-BERT, which indicates both the effectiveness of the visual-linguistic pre-training framework, and the incorporation of continuous (captions) and discrete (answer categories) latent variables.

Compared to the results on the standard baseline (UpDn), the improvements achieved by our proposed model on the VL-BERT framework is smaller. 
This is because VL-BERT has been pre-trained on massive image captioning data, where the learning of visual features have largely benefited from the modality of captions already.
Nevertheless, based upon the strong baseline model, our proposed model can still improve performance slightly, which further indicates the effectiveness of the latent variable framework. 

The state-of-the-art performance on VQA v2.0 among pre-training frameworks is achieved by LXMERT, Oscar and Uniter \cite{tan-bansal-2019-lxmert, li2020oscar, chen2020uniter}.
They have been extensively pre-trained using massive datasets on languages and vision tasks (including VQA) in a multi-task learning fashion. 
Our work is not directly comparable, and is not aimed at improving and beating the state-of-the-art performance. 
Instead, it is focused on exploring the potential of latent variable models to represent additional useful information in multimodal learning and to contribute to pre-trained vision-language frameworks.
We draw attention to the advantage of using generative framework on the VQA task. 
In this case, we can employ more information during training (which is omitted in testing) to regularise the original multimodal distribution.
This can be demonstrated by the improvements on VL-BERT brought by the latent variables.


\begin{table}[t]
\centering
\addtolength{\tabcolsep}{2pt}
\scalebox{0.95}{
\begin{tabular}{l|cccc}
\hline 
\toprule
& \multicolumn{4}{|c}{VQA v2.0 val} \\
\hline
& {\footnotesize All} & {\footnotesize Yes/No} & {\footnotesize Num} &  {\footnotesize Other} \\
\hline
\hline
UpDn & 63.15 & 80.38 & 42.84 & 55.86 \\
UpDn + caption & 63.85 & 81.10 & 43.63 & 55.90 \\
UpDn + category & 63.51 & 81.62 & 42.17 & 55.38 \\
\midrule
Ours w/o caption &  \bf 64.09 &  81.82 &  44.37 &  55.74 \\
Ours w/ caption &  \underline{64.24} &  82.36 &  44.52 &  56.02 \\
\bottomrule
\end{tabular}}
\caption{Ablation study to investigate the effect of each component: caption, and answer category. ``Ours w/o caption" indicates our final model in which only image and question pairs are needed at test time; while ``Ours w/ caption" represents the model using caption during evaluation. The result of our best model are in bold; while the best performance with captions as inputs during testing is underlined for comparison.}
\label{tab:ablation}
\end{table}

\subsection{Qualitative Analysis}
We perform an ablation study to qualitatively analyse the effect of the components introduced in our work brought by the continuous (image caption) and discrete (answer category) latent variables, as shown in Table~\ref{tab:ablation}.

\subsubsection{Effect of Captions}
The introduction of captions as a continuous latent variable improves the classification performance, with an additional modality as input to benefit the learning of multimodal representations.
According to the breakdown numbers in ~\ref{tab:ablation},
the improvements brought by the latent variables of captions and answer categories are 0.70 and 0.36 respectively for All questions altogether. 
The combined strategy reaches 0.94 which indicates that the benefits from the two latent variables are complementary.
Note that neither the captions nor the answer categories is available during testing; 
we only make use of these modalities during training.

To further investigate potential benefit of the captions, we design an experiment that feed in ground truth captions via variational distribution for caption representations instead of inferring them from question and answer pairs (\ie use $q_{\phi}(z|c)$ to replace $p_{\theta}(z|v,q)$). 
We test this out in the validation dataset and obtain 64.24 (`Ours w/ caption') compared to 64.09 (`Ours w/o caption').
It shows that having explicit captions as input gives slightly better performance.
However, the captions in these experiments are ground truth, which means that if we were to use instead automatically generated captions from an image captioning pipeline, the numbers might drop due to the possible captioning errors. 
Primarily, our proposed model (`Ours w/o caption') achieves the performance on par with with the model with ground truth captions, which demonstrates the effectiveness of the strategy that incorporates extra modality by latent variables. 

\begin{figure}[t]
  \centering
  \includegraphics[width=0.5\textwidth]{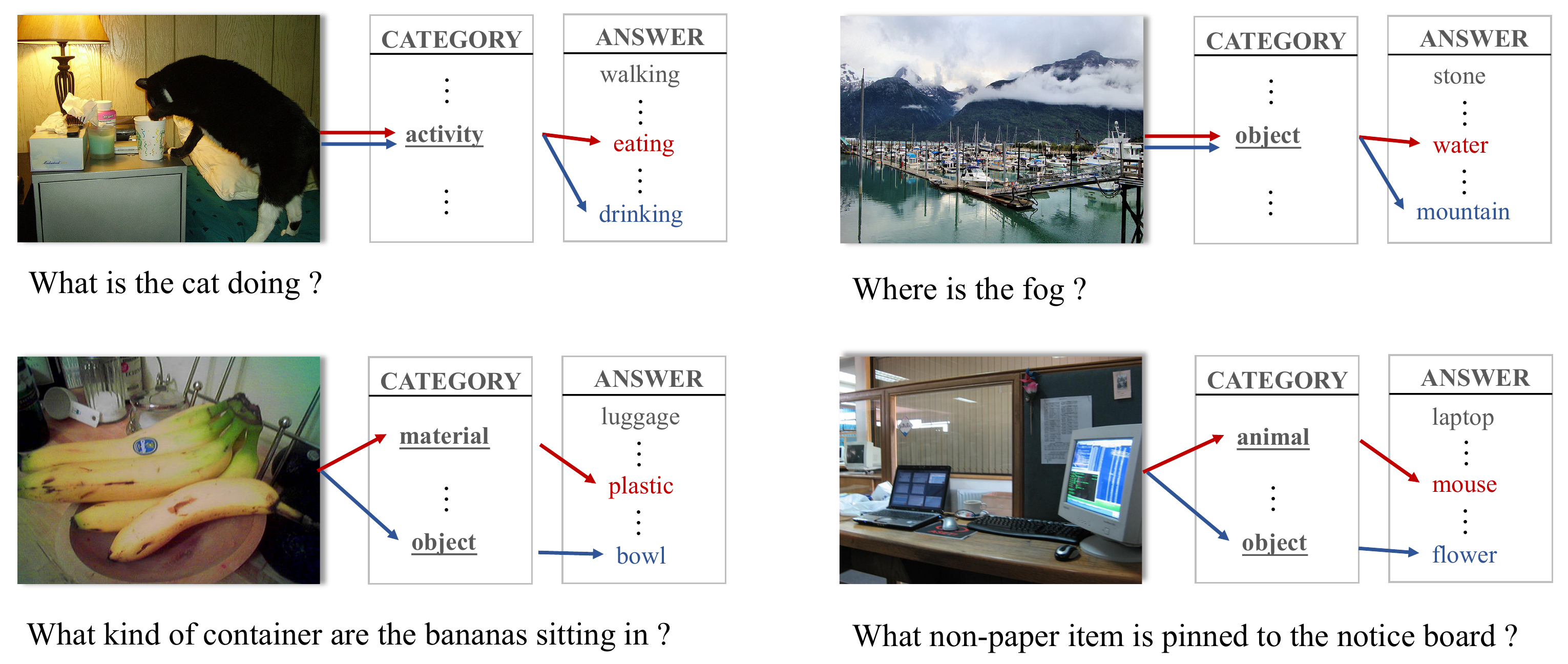}
  \caption{Examples of our latent variable model outperforming the baseline UpDn model from the introduction of answer category as a discrete latent variable. The answer predicted by UpDn is highlighted in red and the answer from our model is in blue. We also show other sample answer candidates within each category.}
  \label{fig:ablation}
\end{figure}

\subsubsection{Effect of Answer Category}
As it can be observed from Table~\ref{tab:ablation}, after introducing answer category as an additional discrete latent variable, our proposed model can also be improved over the UpDn baseline, where the largest improvement can be observed for the ``Yes/No" type. 
For the question types ``Num'' and ``Other'', the results of \texttt{UpDn+category} are lower than the baseline.
This may be due to the multiple answer candidates under the two categories. 
For example, although the category classier can accurately predict the answer category (\eg, ``count'', ``color'', \etc), it can be still difficult to distinguish among the answers - \{``9'', ``20'', ..., ``many'' for ``count'';  ``black'', ``brown'', ..., ``black and white'' for ``color''\}.
We highlight that the contribution of answer categories as a discrete latent variable is to introduce an inductive bias which helps predict the correct answer categories and answers given a specific image and question.

In order to further elaborate the effectiveness of answer categories, we extract examples where our model predicted the correct answers while the UpDn baseline failed to do so, as shown in Figure~\ref{fig:ablation}.
For the two cases in the top row, both models predict answers under the same and correct answer categories, hence  the answer space is similar; however, our latent variable model can effectively distinguish and learn the difference among the answers which fall within the same category. 
The bottom row of Figure~\ref{fig:ablation} shows  two cases where the two models predict answers in different answer categories, and therefore they are also very different in meaning. 
Our model not only outputs the highest probability for the correct answer category, but also makes the correct final prediction. 

\section{Conclusions}
In this paper, we propose to tackle VQA under the framework of latent variable models, employing captions and answer categories as the continuous and the discrete latent variables respectively to constrain the original image-question distribution while omitting the extra information during the test phase.
Our experimental results and qualitative analysis show the effectiveness of the latent variables in boosting answering performance at test time when only image and question pairs are available. 
This framework could be easily generalised to incorporate other types of information or modalities to enhance VQA and other tasks. 


{\small
\bibliographystyle{ieee_fullname}
\bibliography{egbib}
}

\end{document}